\documentclass[letterpaper]{article}
\usepackage{aaai}
\usepackage{times}
\usepackage{latexsym}
\usepackage{amsmath}
\usepackage{multirow}
\usepackage{url}
\usepackage{graphicx}
\usepackage{amssymb}
\usepackage{epsfig}
\usepackage{verbatim}
\usepackage{caption}

\newtheorem{definition}{Definition}

\title{Solving puzzles described in English by automated translation to answer set programming and learning how to do that translation}

\author{Chitta Baral\\
  School of Computing, Informatics and DSE\\
    Arizona State University\\ 
  \texttt{chitta@asu.edu} \\\And
 Juraj Dzifcak\\
 School of Computing, Informatics and DSE\\
Arizona State University\\ 
       \texttt{juraj.dzifcak@asu.edu} }

\date{}
\begin{document}
\maketitle

\begin{abstract}
We present a system capable of automatically solving combinatorial logic puzzles given in (simplified) English. It involves translating the English descriptions of the puzzles into answer set programming(ASP) and using ASP solvers to provide solutions of the puzzles. To translate the descriptions, we use a $\lambda$-calculus based approach using Probabilistic Combinatorial Categorial Grammars (PCCG) where the meanings of words are associated with parameters to be able to distinguish between multiple meanings of the same word.  Meaning of many words and the parameters are learned. The puzzles are represented in ASP using an ontology which is applicable to a large set of logic puzzles. 
\end{abstract}

\section{Introduction and Motivation}

Consider building a system that can take as input an English description of combinatorial logic puzzles\footnote{An example is the well-known Zebra puzzle. 
http://en.wikipedia.org/wiki/Zebra\_Puzzle}  \cite{puzzles} and solve those puzzles. Such a system would need and somewhat demonstrate the ability to (a) process language, (b) capture the knowledge in the text and (c) reason and do problem solving by searching over a space of possible solutions.  Now if we were to build this system using a larger system that learns how to process new words and phrases then the latter system would need and somewhat demonstrate the ability of (structural) learning. The significance of the second larger system is with respect to being able to learn language (new words and phrases) and not expecting that humans will a-priori provide an exhaustive vocabulary of all the words and their meanings.

In this paper we describe our development of such a system with {\em some added assumptions.} We present evaluation of our system in terms of how well it learns to understand clues (given in simplified\footnote{Our simplified English is different from ``controlled'' English in that it does not have a pre-specified grammar. We only do some preprocessing to eliminate anaphoras and some other aspects.} English) of puzzles and how well it can solve new puzzles. Our approach of solving puzzles given in English involves translating the English description of the puzzles to sentences in answer set programming (ASP) \cite{baral:book} and then using ASP solvers, such as \cite{clingo}, to solve the puzzles. Thus a key step in this is to be able to translate English sentences to ASP rules. A second key step is to come up with an appropriate ontology of puzzle representation that makes it easy to do the translation.

With respect to the first key step, we use a methodology \cite{me:iwcs} that assigns $\lambda$-ASP-Calculus\footnote{$\lambda$-ASP-Calculus is inspired by $\lambda$-Calculus. The classical logic formulas in $\lambda$-Calculus are replaced by ASP rules in $\lambda$-ASP-Calculus.} rules to each words. Since it seems to us that it is not humanly possible to manually create $\lambda$-ASP-Calculus rules for English words, we have developed a method, which we call, Inverse $\lambda$ to learn the meaning of English words in terms of their $\lambda$-ASP-Calculus rule. The overall architecture of our system is given in Figure 1. Our translation (from English to ASP) system, given in the left hand side of Figure 1, uses a Probabilistic Combinatorial Categorial Grammars (PCCG) \cite{Mooney:2005} and a lexicon consisting of words, their corresponding $\lambda$-ASP-Calculus rules and associated (quantitative) parameters to do the translation. Since a word may have multiple meaning implying that it may have multiple associated $\lambda$-ASP-Calculus rules, the associated parameters help us in using the ``right'' meaning in that the translation that has the highest associated probability is the one that is picked.  Given a training set of sentences and their corresponding $\lambda$-ASP-Calculus rules, and an initial vocabulary (consisting of some words and their meaning), Inverse $\lambda$ and generalization is used to guess the meaning of words which are encountered but are not in the initial lexicon. Because of this guess and because of inherent ambiguity of words having multiple meanings, one ends up with a lexicon where words are associated with multiple $\lambda$-ASP-Calculus rules. A parameter learning method is used to assign weights to each meaning of a word in such a way that the probability that each sentence in the training set would be translated to the given corresponding $\lambda$-ASP-Calculus rule is maximized. The block diagram of this learning system is given in the right hand side of Figure 1.

\begin{figure*}[!ht]
  \begin{center}
      \includegraphics[width=12cm]{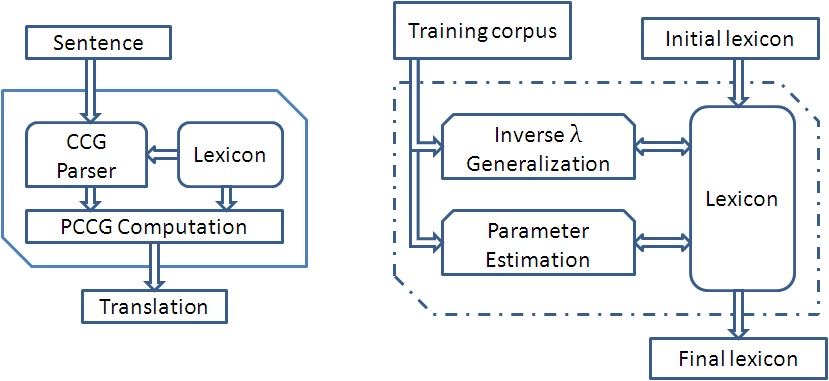}
      \caption{Overall system architecture}
      \label{figarch}
   \end{center}
\end{figure*}

With respect to the second key step, there are many ASP encodings, such as in \cite{baral:book}, of combinatorial logic puzzles. However, most methods given in the literature, assume that a human is reading the English description of the puzzle and is coming up with the ASP code or code in some high level language \cite{Mirek02} that gets translated to ASP. In our case the translation of English description of the puzzles to ASP is to be done by an automated system and moreover this systems learns aspects of the translation by going over a training set. This means we need an ontology of how the puzzles are to be represented in ASP that is applicable to most (if not all) combinatorial logic puzzles.  

The rest of the paper is organized as follows: We start by discussing the assumptions we made for our system. We then provide an overview of the ontology we used to represent the puzzles. We then give an overview of the natural language translation algorithm followed by a simple illustration on a small set of clues. Finally, we provide an evaluation of our approach with respect to translating clues as well as translating whole puzzles. We then conclude.

\section{Assumptions and Background Knowledge}

With our longer term goal to be able to solve combinatorial logic puzzles specified in English, as mentioned earlier, we made some simplifying assumptions for this current work. Here we assumed that the domains of puzzles are given (and one does not have to extract it from the puzzle description) and focused on accurately translating the clues. Even then English throws up many challenges and we did a human preprocessing\footnote{The people doing the pre-processing were not told of any specific subset of English or any ``Controlled'' English to use. They were only asked to simplify the sentences so that each sentence would translate to a single clue.} of puzzles to eliminate anaphoras and features that may lead to a sentence being translated into multiple clues. Besides translating the given English sentences we added some domain knowledge related to combinatorial logic puzzles. This is in line with the fact that often natural language understanding involves going beyond literal understanding of a given text and taking into context some background knowledge. The following example illustrates these points. A clue ``Earl arrived immediately before the person with the Rooster.'' specifies several things. Outside of the fact that a man with the first name ``Earl'' came immediately before the man with the animal ``Rooster'', a human would also immediately conclude that ``Earl'' does not have a ``Rooster''. To correctly process this information one needs the general knowledge that if person $A$ arrives before person $B$, $A$ and $B$ are different persons and given the assumption that all the objects are exclusive, an animal has a single owner.  Also, to make sure that clue sentences correspond to single ASP rules, during preprocessing of this clue one may add  ``Earl is not the person with the Rooster.''

\section{Puzzle representation and Ontology}

For our experiments, we focus on logic puzzles from \cite{puzzles,puzzles1,puzzles2}. These logic puzzles have a set of basic domain data and a set of clues. To solve them, we adopt an approach where all the possible solutions are generated, and then constraints are added to reduce the number of solutions. In most cases there is a unique solution.  A sample puzzle is given below, whose solution involves finding the correct  associations between persons, their ranks, their animals and  their lucky elements.

\begin{small}
\begin{verbatim}
Puzzle Domain data:
1,2,3,4 and 5 are ranks
earl, ivana, lucy, philip and tony are names
earth, fire, metal, water and wood are elements
cow, dragon, horse, ox and rooster are animals

Puzzle clues:

1) Tony was the third person to have his
fortune told.
2) The person with the Lucky Element Wood 
had their fortune told fifth.
3) Earl's lucky element is Fire.
4) Earl arrived immediately before the 
person with the Rooster.
5) The person with the Dragon had their 
fortune told fourth.
6) The person with the Ox had their 
fortune told before the one 
who's Lucky Element is Metal.
7) Ivana's Lucky Animal is the Horse.
8) The person with the Lucky Element 
Water has the Cow.
9) The person with Lucky Element Water 
did not have their fortune told first.
10) The person with Lucky Element Earth 
had their fortune told exactly 
two days after Philip.
\end{verbatim}
\end{small}

The above puzzle can be encoded as follows.

\begin{small}
\begin{verbatim}

% DOMAIN DATA

index(1..4).
eindex(1..5).

etype(1, name). 
element(1,earl). element(1,ivana). 
element(1,lucy). element(1,philip). 
element(1,tony).
etype(2, element). 
element(2,earth). element(2,fire). 
element(2,metal). element(2,water). 
element(2,wood).
etype(3, animal). 
element(3,cow). element(3,dragon). 
element(3,horse). element(3,ox). 
element(3,rooster).
etype(4, rank). 
element(4,1). element(4,2). element(4,3). 
element(4,4). element(4,5).

% CLUES and their translation

%Tony was the third person to have 
%his fortune told.
:- tuple(I, tony), tuple(J, 3), I!=J.

%The person with the Lucky Element 
%Wood had their fortune told fifth.
:- tuple(I, wood), tuple(J, 5), I!=J.

%Earl's lucky element is Fire.
:- tuple(I, earl), tuple(J, fire), I!=J.

%Earl arrived immediately before 
%the person with the Rooster.
:- tuple(I, earl), tuple(J, rooster), 
tuple(I, X), tuple(J, Y), 
etype(A, rank), element(A, X), 
element(A, Y), X != Y-1.

%The person with the Dragon had 
%their fortune told fourth.
:- tuple(I, dragon), tuple(J, 4), I!=J.

%The person with the Ox had their 
% fortune told before the
%one who's Lucky Element is Metal.
:- tuple(I, ox), tuple(J, metal), 
tuple(I, X), tuple(J, Y), 
etype(A, rank), element(A, X), 
element(A, Y), X > Y.

%Ivana's Lucky Animal is the Horse.
:- tuple(I, ivana), tuple(J, horse), I!=J.

%The person with the Lucky Element 
%Water has the Cow.
:- tuple(I, water), tuple(J, cow), I!=J.

%The person with Lucky Element Water 
%did not have their fortune told first.
:- tuple(I, water), tuple(I, 1).

%The person with Lucky Element Earth 
%had their fortune 
%told exactly two days after  Philip.
:- tuple(I, earth), tuple(J, philip), 
tuple(I, X), tuple(J, Y), 
etype(A, rank), element(A, X), 
element(A, Y), X != Y+2.
\end{verbatim}
\end{small}

\subsection{The puzzle domain data}

Each puzzle comes with a set of basic domain data which forms tuples. An example of this data is given above. Note that this is not the format in which they are provided in the actual puzzles. It is assumed that the associations are exclusive, e.g. ``earl'' can own either a ``dragon'' or a ``horse'', but not both. We assume this data is provided as input. There are several reasons for this assumption. The major reason is that not all the data is given in the actual natural language text describing the puzzle. In addition, the text does not associate actual elements, such as ``earth'' with element types, such as ``element''. If the text contains the number ``6'', we might assume it is a rank, which, in fact, it is not. These domain data is encoded using the following format, where $etype(A,t)$ stores the element type $t$, while $element(A,X)$ is the predicate storing all the elements $X$ of the type $etype(A,type)$. An example of an instance of this encoding is given below.

\begin{small}
\begin{verbatim}
% size of a tuple
index(1..n).
% number of tuples
eindex(1..m).

% type and lists of elements of that type, 
% one element from 
% each index forms a tuple
etype(1, type1).
element(1, el11). element(1, el12). ...  
element(1, el1n).
...
etype(m, typem).
element(m, em11). element(1, elm2). ...  
element(1, elmn).
\end{verbatim}
\end{small}

We now discuss this encoding in more detail. We want to encode all the elements of a particular type, The type is needed in order to do direct comparisons between the elements of some type. For example, when we want to specify that ``Earl arrived immediately before the person with the Rooster.'', as encoded in the sample puzzle, we want to encode something like  $etype(A, rank), element(A, X), element(A, Y), X != Y-1.$, which compares the ranks of elements $X$ and $Y$. The reason all the element types and elements have fixed numerical indices is to keep the encoding similar across the board and to not have to define additional grounding for the variables. For example, if we encoded elements as $element(name,earl)$, then if we wanted to use the variable $A$ in the encodings of the clue, it would have to have defined domain which includes all the element types. These differ from puzzle to puzzle, and as such would have to be specifically added for each puzzle. By using the numerical indices across all puzzles, these are common across the board and we just need to specify that $A$ is an index. In addition, to avoid permutation within the tuples, the following facts are generated, where $tuple(I,X)$ is the predicate storing the elements $X$ within a tuple $I$:

\begin{small}
\begin{verbatim}
tuple(1,e11). ... tuple(1,e1n).
\end{verbatim}
\end{small}

which for the particular puzzle yields

\begin{small}
\begin{verbatim}
tuple(1, 1). tuple(2, 2). tuple(3, 3). 
tuple(4, 4).tuple(5,5).
\end{verbatim}
\end{small}

\subsection{Generic modules and background knowledge}

Given the puzzle domain data, we combine their encodings with additional modules responsible for generation and generic knowledge. In this work, we assume there are two type of generic modules available. The first one is responsible for generating all the possible solutions to the puzzle. We assume these are then pruned by the actual clues, which impose constraints on these. The following rules are responsible for generation of all the possible tuples. Recall that we assume that all the elements are exclusive. 

\begin{small}
\begin{verbatim}
1{tuple(I,X):element(A,X)}1.
:- tuple(I,X), tuple(J,X), 
element(K,X), I != J.
\end{verbatim}
\end{small}

% maximum, minimum, etc.
In addition, a module with rules defining generic/background knowledge is used so as to provide higher level knowledge which the clues define. For example, a clue might discuss maximum, minimum, or genders such as woman. To be able to match these with the puzzle data, a set of generic rules defining these concepts is used, rather than adding them into the actual puzzle data. Thus rules defining concepts and knowledge such as maximum, minimum, within range, sister is a woman and others are added. For example, the concept ``maximum'' is encoded as:

\begin{small}
\begin{verbatim}
notmax(A, X) :- element(A, X), 
              element(A, Y), X != Y, Y > X.
maximum(A, X) :- not notmax(A,X), 
                 element(A,X).
\end{verbatim}
\end{small}

%We assume the presented natural language approach can be learned to obtain these concepts as well.

\subsection{Extracting relevant facts from the puzzle clues}

A sample of clues with their corresponding representations is given in the sample puzzle above. Let us take a closer look at the clue ``Tony was the third person to have his fortune told.'', encoded as $:- tuple(I, tony), tuple(J, 3), I\neq J$. This encoding specifies that if ``Tony'' is assigned to tuple $I$, while the rank ``3'' is assigned to a different tuple $J$, we obtain false. Thus this ASP rule limits all the models of it's program to have ``Tony'' assigned to the same tuple as ``3''. One of the questions one might ask is where are the semantic data for ``person'' or ``fortune told''. They are missing from the translation since with respect to the actual goal of solving the puzzle, they do not contribute anything meaningful. The fact that ``Tony'' is a ``person'' is inconsequential with respect to the solutions of the puzzle. With this encoding, we attempt to encode only the relevant information with regards to the solutions of the puzzle. This is to keep the structure of the encodings as simple and as general as possible. In addition, if the rule would be encoded as $:- person(tony), tuple(I, tony), tuple(J, 3), I\neq J.$, the fact $person(tony)$ would have to be added to the program in order for the constraint to give it's desired meaning. However, this does not seem reasonable as there are no reasons to add it (outside for the clue to actually work), since ``person'' is not present in the actual data of the puzzle.

\section{Translating Natural language to ASP}
%
%
%Several sample clues with their corresponding translations are given below.
%
%\begin{itemize}
%\item ``Martin does not have surname Iglesias.''
%
%$:- tuple(I,martin), tuple(I,iglesias).$
%
%\item ``The man known as fierce is on the Tower's fourth floor.''
%
%$:- tuple(I, fierce), tuple(J,4), I \neq J.$
%
%\item ``Earl arrived immediately before the person with the Rooster''.
%
%$:- tuple(I, earl), tuple(J, rooster),tuple(I, X), $
%
%$tuple(J, Y), etype(A, time), $
%
%$element(A, X), element(A, Y), X != Y-1.$
%
%\item ``Kevin came before the man with surname Han.''
%
%$:- tuple(I,kevin), tuple(J,han), tuple(I,X), tuple(J,Y),$ 
%
%$etype(A,rank), element(A,X), element(A,Y),$ 
%
%$X \geq Y. $
%
%\item ``If a skier did a Nordic event, the skier did all Nordic events.''
%
%$:- tuple(I,X), not tuple(I,Y), etype(X,event),$ 
%
%$etype(Y,event), nordic(X), nordic(Y),$
%
%$X != Y.$
%
%\item ``A skier did not do both a Nordic event and an Alpine event.''
%
%$:- tuple(I,X), tuple(I,Y), etype(X,event), $
%
%$etype(Y,event), alpine(X), nordic(Y).$
%
%\end{itemize}

To translate the english descriptions into ASP, we adopt our approach in \cite{me:iwcs}. This approach uses $inverse$-lambda computations, generalization on demand and trivial semantic solutions together with learning. However for this paper, we had to adapt the approach to the ASP language and develop an ASP-$\lambda$-Calculus.  An example of a clue translation using combinatorial categorial grammar \cite{Steedman:Book} and ASP-$\lambda$-calculus is given in table \ref{tab:ex1}.

%The overall system architecture is given by figure \ref{figarch}. To the left we present an overall system for translating a sentence into a logical form, while on the right there are training modules used to train the dictionary containing the semantic representations of words. 

%\begin{figure*}[!ht]
%  \begin{center}
%      \includegraphics[width=12cm]{outline}
%      \caption{Overall system architecture}
%      \label{figarch}
%   \end{center}
%\end{figure*}

The system uses the two inverse $\lambda$ operators, $Inverse_L$ and $Inverse_R$ as given in \cite{me:iwcs} and \cite{Marcos:thesis}. Given $\lambda$-calculus formulas $H$ and $G$, these allow us to compute a $\lambda$-calculus formula $F$ such that $H = F @ G$ and $H = G @ F$. We now present one of the two Inverse $\lambda$ operators, $Inverse_R$ as given in \cite{me:iwcs}. For more details, as well as the other operator, please see \cite{Marcos:thesis}.We now introduce the different symbols used in the algorithm and their meaning :

\begin{itemize}
\item Let $G$, $H$ represent typed $\lambda$-calculus formulas, $J^1$,$J^2$,...,$J^n$ represent typed terms, $v_1$ to $v_n$, $v$ and $w$ represent variables and $\sigma_1$,...,$\sigma_n$ represent typed atomic terms.
\item Let $f()$ represent a typed atomic formula. Atomic formulas may have a different arity than the one specified and still satisfy the conditions of the algorithm if they contain the necessary typed atomic terms.
\item Typed terms that are sub terms of a typed term J are denoted as $J_i$.
\item If the formulas we are processing within the algorithm do not satisfy any of the $if$ conditions then the algorithm returns $null$.
\end{itemize}

\begin{definition}[operator :]
Consider two lists of typed $\lambda$-elements A and B, $(a_i,...,a_n)$ and $(b_j,...,b_n)$ respectively and a formula $H$. The result of the operation $H(A:B)$  is obtained by replacing $a_i$ by $b_i$, for each appearance of A in H.
\end{definition}

Next, we present the definition of an inverse operators\footnote{This is the operator that was used in this implementation. In a companion work we develop an enhancement of this operator which is proven sound and complete.} $Inverse_R(H,G)$: \\

\begin{definition}%[$Inverse_R(H,G)$]

The function $Inverse_R(H,G)$ is defined as:\newline
\noindent Given $G$ and $H$:
\begin{it}
\begin{enumerate}
\item If $G$ is $\lambda v.v@J$, set $F = Inverse_L(H,J)$
\item If $J$ is a sub term of $H$ and G is $\lambda v.H(J:v)$
\begin{itemize}
\item  $F$ = $J$
\end{itemize}
\item G is not $\lambda v.v@J$, $J$ is a sub term of $H$ and G is $\lambda w.H(J(J_1,...,J_m):w@J_p,...,@J_q)$ with 1 $\leq$ p,q,s $\leq$ m.
\begin{itemize}
\item $F$ = $\lambda v_1,...,v_s.J(J_1,...,J_m:v_p,...,v_q)$.
\end{itemize}

\end{enumerate}
\end{it}
\end{definition}

Lets assume that in the example given by table \ref{tab:ex1} the semantics of the word ``immediately'' is not known. We can use the Inverse operators to obtain it as follows. Using the semantic representation of the whole sentence as given by table \ref{tab:ex1}, and the word ``Earl'',$\lambda x. tuple(x,earl)$, we can use the respective operators to obtain the semantic of ``arrived immediately before the man with the Rooster'' as 
\begin{small}
$\lambda z. :- z@I, tuple(J,rooster), tuple(I,X),$ $ tuple(J,Y), etype(A,rank), element(A,X), element(A,Y),$ $X \neq Y-1.$
\end{small}

Repeating this process recursively we obtain $\lambda x. \lambda y. x \neq y-1$ as the representation of ``arrived immediately'' and $\lambda x. \lambda y. \lambda z. x @ (y \neq z-1)$ as the desired semantic for ``immediately''.

The input to the overall learning algorithm is a set of pairs $(S_i,L_i), i=1,...,n$, where $S_i$ is a sentence and $L_i$ its corresponding logical form. The output of the algorithm is a PCCG defined by the lexicon $L_T$ and a parameter vector $\Theta_T$. As given by \cite{me:iwcs}, the parameter vector $\Theta_i$ is updated at each iteration of the algorithm. It stores a real number for each item in the dictionary. The overall learning algorithm is given as follows:

\begin{small}
\begin{itemize}
\item {\bf Input:} A set of training sentences with their corresponding desired representations $S = \{(S_i,L_i) : i = 1 . . . n\}$ where
$S_i$ are sentences and $L_i$ are desired expressions. Weights are given an initial value of $0.1$.

An initial feature vector $\Theta_0$.
An initial lexicon $L_0$.

\item {\bf Output}: An updated lexicon $L_{T+1}$. 
An updated feature vector $\Theta_{T+1}$.

\item {\bf Algorithm:}

\begin{itemize}

\item Set $L_0$

\item For t = 1 . . . T

\item Step 1: (Lexical generation)

\item For i = 1...n.
\begin{itemize}
\item For j = 1...n.

\item Parse sentence $S_j$ to obtain $T_j$
\item Traverse $T_j$
\begin{itemize}
\item  apply $INVERSE\_L$, $INVERSE\_R$ and $GENERALIZE_D$ to find new $\lambda$-calculus expressions of words and phrases $\alpha$.
\end{itemize}

\item Set $L_{t+1} = L_t \cup \alpha$

\end{itemize}
\item Step 2: (Parameter Estimation)

\item Set $\Theta_{t+1} = UPDATE(\Theta_t, L_{t+1})$\footnote{For details on $\Theta$ computation, please see \cite{Collins:2005} }
\end{itemize}
\item return $GENERALIZE(L_T, L_T), \Theta(T)$ 
\end{itemize}
\end{small}

To translate the clues, a trained model was used to translate these from natural language into ASP. This model includes a dictionary with $\lambda$-calculus formulas corresponding to the semantic representations of words. These have their corresponding weights.

\begin{table*}[htb]
\tiny{
\begin{center}
\begin{tabular}{c c c c c c c c c}
Earl & arrived & immediately & before & the & man & with & the & Rooster. \\
$NP$ & $S \backslash NP$ & $(S \backslash NP) \backslash (S \backslash NP)$ & $((S \backslash NP) \backslash (S \backslash NP)) / NP$ &  $NP/N$ & $N$ & $(NP \backslash NP) / NP$ & $NP/N$ & $N$\\
\cline{2-3}\cline{5-6}\cline{8-9}
$NP$ & & $S \backslash NP$ & $((S \backslash NP) \backslash (S \backslash NP)) / NP$ &  $NP$ & & $(NP \backslash NP) / NP$ & $NP$ & \\
\cline{2-3}\cline{5-6}\cline{7-9}
$NP$ & & $S \backslash NP$ & $((S \backslash NP) \backslash (S \backslash NP)) / NP$ &  $NP$ & & $NP \backslash NP$ & & \\
\cline{2-3}\cline{5-9}
$NP$ & & $S \backslash NP$ & $((S \backslash NP) \backslash (S \backslash NP)) / NP$ &  & & $NP$ & & \\
\cline{2-3}\cline{4-9}
$NP$ & & $S \backslash NP$ & $(S \backslash NP) \backslash (S \backslash NP)$ &  & & & & \\
\cline{2-9}
$NP$ & & & $(S \backslash NP)$ &  & & & & \\
\cline{1-9}
 & & & $S$ &  & & & & \\
\end{tabular}

\begin{tabular}{c c c}
earl & arrived & immediately \\
$\lambda x. tuple(x,earl)$ & $\lambda x. x$ & $\lambda x. \lambda y. \lambda z. x @ (y \neq z-1)$\\
\cline{2-3}
%\cline{4-5}\cline{7-8}
$\lambda x. tuple(x,earl)$ & & $\lambda x. \lambda y. x \neq y-1$ \\
\cline{2-3}
%\cline{4-5}\cline{6-8}
$\lambda x. tuple(x,earl)$ & & $\lambda x. \lambda y. x \neq y-1$ \\
\cline{2-3}
%\cline{4-8}
$\lambda x. tuple(x,earl)$ & & $\lambda x. \lambda y. x \neq y-1$ \\
\cline{2-3}
%\cline{3-8}
$\lambda x. tuple(x,earl)$ & & $\lambda x. \lambda y. x \neq y-1$ \\
\cline{2-3}
$\lambda x. tuple(x,earl)$ & & \\
\cline{1-3}
 & & \\
\end{tabular}

\begin{tabular}{c} 
before \\
 $\lambda x. \lambda y. \lambda z. :- z@I, x@J, tuple(I,X), tuple(J,Y), etype(A,rank), element(A,X), element(A,Y), y@X@Y.$ \\
 $\lambda x. \lambda y. \lambda z. :- z@I, x@J, tuple(I,X), tuple(J,Y), etype(A,rank), element(A,X), element(A,Y), y@X@Y.$ \\
 $\lambda x. \lambda y. \lambda z. :- z@I, x@J, tuple(I,X), tuple(J,Y), etype(A,rank), element(A,X), element(A,Y), y@X@Y.$ \\
 $\lambda x. \lambda y. \lambda z. :- z@I, x@J, tuple(I,X), tuple(J,Y), etype(A,rank), element(A,X), element(A,Y), y@X@Y.$ \\
 $\lambda y. \lambda z. :- z@I, tuple(J,rooster), tuple(I,X), tuple(J,Y), etype(A,rank), element(A,X), element(A,Y), y@X@Y.$ \\
 $\lambda z. :- z@I, tuple(J,rooster), tuple(I,X), tuple(J,Y), etype(A,rank), element(A,X), element(A,Y), X \neq Y-1.$ \\
 $:- tuple(I,earl), tuple(J,rooster), tuple(I,X), tuple(J,Y), etype(A,rank), element(A,X), element(A,Y), X \neq Y-1.$ \\
\end{tabular}

\begin{tabular}{c c c c c} 
the & man & with & the & Rooster. \\
$\lambda x. x$ & $\lambda x. x$ & $\lambda x. \lambda y. y@x$ & $\lambda x. x$ & $\lambda x. tuple(x,rooster)$\\
\cline{1-2}\cline{4-5}
$\lambda x. x$ & & $\lambda x. \lambda y. y@x$ & $\lambda x. tuple(x,rooster)$ & \\
\cline{1-2}\cline{3-5}
$\lambda x. x$ & & $\lambda x. \lambda y. y@(\lambda x. tuple(x,rooster))$ & & \\
\cline{1-5}
 & & $\lambda x. tuple(x,rooster)$ & & \\

%\cline{2-8}
% & & & & \\
%\cline{1-8}
% & & & & \\
\end{tabular}

\end{center}
}
\caption{CCG and $\lambda$-calculus derivation for ``Earl arrived immediately before the person with the Rooster.''}
\label{tab:ex1}
\end{table*}

\begin{table*}[htb]
\tiny{
\begin{center}
\begin{tabular}{c c c c c}
Miss Hanson & is & withdrawing & more & than the customer whose number is 3989. \\
$NP$ & $(S/NP) \backslash NP$ & $(S\backslash (S/NP))/NP$& $NP/NP$ & $NP$ \\
\cline{1-2}\cline{4-5}
 & $S/NP$ & $(S\backslash (S/NP))/NP$ & $NP$ & \\
\cline{1-2}\cline{3-5}
 & $S/NP$ & $(S\backslash (S/NP))$ & & \\
\cline{1-5}
 & & $S$ & & \\
\end{tabular}

\begin{tabular}{c c c}
Miss Hanson & is & withdrawing \\
$\lambda x. tuple(x,hanson)$ & $\lambda x. \lambda y. (y@x).$ & $\lambda x. \lambda z. (x@z)$  \\
\cline{1-2}
 & $\lambda y. (y@(\lambda x. tuple(x,hanson))).$ & $\lambda x. \lambda z. (x@z)$\\
\cline{1-2}
 & $\lambda y. (y@(\lambda x. tuple(x,hanson))).$ & \\
\cline{1-3}
 & & \\
\end{tabular}

\begin{tabular}{c}
more \\
$\lambda x. \lambda y. :- y@I, x@J, tuple(I,X), tuple(J,Y), etype(A,rank), element(A,X), element(A,Y), X > Y, I!=J.$\\
\cline{1-1}
$\lambda y. :- y@I, tuple(J,3989), tuple(I,X), tuple(J,Y), etype(A,rank), element(A,X), element(A,Y), X > Y, I!=J.$\\
\cline{1-1}
$\lambda z. :- z@I, tuple(J,3989), tuple(I,X), tuple(J,Y), etype(A,rank), element(A,X), element(A,Y), X > Y, I!=J.$ \\
\cline{1-1}
 $:- tuple(I, hanson), tuple(J,3989), tuple(I,X), tuple(J,Y), etype(A,rank), element(A,X), element(A,Y), X > Y, I!=J.$ \\
\end{tabular}

\begin{tabular}{c}
than the customer whose number is 3989. \\
$\lambda x. tuple(x,3989)$ \\
\cline{1-1}
\\
\cline{1-1}
\\
\cline{1-1}
\\
\end{tabular}

\end{center}
}
\caption{CCG and $\lambda$-calculus derivation for ``Miss Hanson is withdrawing more than the customer whose number is 3989.''}
\label{tab:ex2}
\end{table*}

Tables \ref{tab:ex1} and \ref{tab:ex2} give two sample translations of a sentence into answer set programming. In the second example, the parse for the ``than the customer whose number is 3989.'' part is not shown to save space. Also note that in general, names and several nouns were preprocessed and treated as a single noun due to parsing issues. The most noticeable fact is the abundance of expressions such as $\lambda x. x$, which basically directs to ignore the word. The main reason for this is the nature of the translation we are performing. In terms of puzzle clues, many of the words do not really contribute anything significant to the actual clue. The important parts are the actual objects, ``Earl'' and ``Rooster'' and their comparison, ``arrived immediately before''. In a sense, the part ``the man with the'' does not provide much semantic contribution with regards to the actual puzzle solution. One of the reasons is the way the actual clue is encoded in ASP. A more complex encoding would mean that more words have significant semantic contributions, however it would also mean that much more background knowledge would be required to solve the puzzles.

\section{Illustration}

We will now illustrate the learning algorithm on a subset of puzzle clues. We will use the following puzzle sentences, as given in table \ref{tab:illSentencesASP}

\begin{table*}[htb]
\tiny{
\begin{center}
\begin{tabular}{|c| c|}
\hline
Donna dale does not have green fleece. & $:-  tuple(I,donna\_dale), tuple(I,green).$\\
\hline
Hy Syles has a brown fleece. & $:- tuple(I,hy\_syles), tuple(J,brown), I!=J.$\\
\hline
Flo Wingbrook's fleece is not red. & $:- tuple(I,flo\_wingbrook), tuple(I,red).$ \\
\hline
Barbie Wyre is dining on hard-boiled eggs. & $:- tuple(I,eggs), tuple(J,barbie\_wyre), I!=J.$ \\
\hline
Dr. Miros altered the earrings. & $:- tuple(I,dr\_miros), tuple(J,earrings), I!=J.$\\
\hline
A garnet was set in Dr. Lukta's piece. & $:- tuple(I,garnet), tuple(J,dr\_lukta))), I!=J.$ \\
\hline
Michelle is not the one liked by 22 & $:- tuple(I, michelle), tuple(I, 22).$ \\
\hline
Miss Hanson is withdrawing more than the customer whose number is 3989. & $:- tuple(I, hanson), tuple(J,3989), tuple(I,X), tuple(J,Y),$\\
&  $etype(A,rank), element(A,X), element(A,Y), X > Y, I!=J.$ \\
\hline
Albert is the most popular. & $:- tuple(I, albert), tuple(J, X), highest(X), I!=J.$\\
\hline
Pete talked about government. & $:- tuple(I, pete), tuple(J, government), I!=J.$ \\
\hline
Jack has a shaved mustache & $:- tuple(I, jack), tuple(J, mustache), I!=J.$ \\
\hline
Jack did not get a haircut at 1 & $:- tuple(I, jack), tuple(I, 1).$ \\
\hline
The first open house was not listed for 100000. & $:- tuple(I, X),first(X), tuple(I, 100000).$ \\
\hline
The candidate surnamed Waring is more popular than the PanGlobal & $:- tuple(I, waring), tuple(J, panglobal), tuple(I, X), tuple(J, Y),$ \\ 
 & $etype(A, time), element(A, X), element(A, Y), X < Y.$ \\
\hline
Rosalyn is not the least popular. & $:- tuple(I, rosalyn), tuple(I, X), lowest(X).$ \\
\hline
\end{tabular}

\end{center}
}
\caption{Illustration sentences for the ASP corpus}
\label{tab:illSentencesASP}
\end{table*}

Lets assume the initial dictionary contains the following semantic entries for words, as given in table \ref{table:illASPdict}. Please note that many of the nouns and noun phrases were preprocessed. 

\begin{table*}[htb]
\tiny{
\begin{center}
\begin{tabular}{|c |c|}
\hline
verb v & $\lambda x. \lambda y. :- y@I, x@J, I!=J.$, $\lambda x. \lambda y. (x@y)$, $\lambda x. \lambda y. (y@x)$ \\
& $\lambda x. x$ \\
\hline
noun n & $\lambda x. tuple(x, n)$, $\lambda x. x$ \\
\hline
noun n with general knowledge & $\lambda x. n(x)$\\
Example:sister, maximum, female,... & \\
\hline
\end{tabular}
\end{center}
}
\caption{Initial dictionary for the ASP corpus}
\label{table:illASPdict}
\end{table*}

The algorithm will than start processing sentences one by one and attempt to learn new semantic information. The algorithm will start with the first sentence, ``Donna dale does not have green fleece.'' Using inverse $\lambda$, the algorithm will find the semantics of ``not'' as $\lambda z. (z @(\lambda x. \lambda y. :- x@I, y@I.)).$. In a similar manner it will continue through the sentences learning new semantics of words. An interesting set of learned semantics as well as weights for words with multiple semantics are given in table \ref{tab:ASPparres}.

\begin{table*}[htb]
\tiny{
\begin{center}
\begin{tabular}{|c|c|c|}
\hline
word & semantics & weight \\
\hline
not & $\lambda z. (z @(\lambda x. \lambda y. :- x@I, y@I.))$ & -0.28 \\
\hline
not & $\lambda y. \lambda x. :- x@I, y@I.)$ & 0.3 \\
\hline
has & $\lambda x. \lambda y. :- y@I, x@J, I!=J.$ & 0.22\\
\hline
has & $\lambda x. \lambda y. (x@y)$ & 0.05\\
\hline
has & $\lambda x. \lambda y. (y@x)$ & 0.05\\
\hline
has & $\lambda x. x$ & 0.05\\
\hline
popular & $\lambda x. tuple(x,popular)$ & 0.17\\
\hline
popular & $\lambda x. x$ & 0.03\\
\hline
a & $\lambda x. x$ & 0.1\\
\hline
not & $\lambda x. \lambda y. :- y@I, x@I.$ & 0.1\\
\hline
on & $\lambda x. x$ & 0.1\\
\hline
the & $\lambda x. x$ & 0.1\\
\hline
in & $\lambda x. \lambda y. (y@x)$ & 0.1\\
\hline
by & $\lambda x. x$ & 0.1\\
\hline
most & $\lambda y. \lambda x. y @ (tuple(x,X), highest(X))$ & 0.1\\
\hline
about & $\lambda x. x$ & 0.1\\
\hline
shaved & $\lambda x. x$ & 0.1\\
\hline
at & $\lambda y. \lambda x. (x@y)$ & 0.1\\
\hline
first & $\lambda y. y @ (\lambda x. tuple(x,X), first(X))$ & 0.1\\
\hline
for & $\lambda x. x$. & 0.1\\
\hline
least & $\lambda x. tuple(x,X), lowest(X)$ & 0.1\\
\hline
more & $\lambda x. \lambda y. :- y@I, x@J, tuple(I,X), tuple(J,Y),$ & \\
 & $etype(A,rank), element(A,X), element(A,Y), X > Y, I!=J.$  & 0.1\\
\hline
\end{tabular}
\end{center}
}
\caption{Learned semantics and final weights of selected words of the ASP corpus.}
\label{tab:ASPparres}
\end{table*}

\section{Evaluation}

We assume each puzzle is a pair $P = (D,C)$ where $D$ corresponds to puzzle domain data, and $C$ correspond to the clues of the puzzle given in simplified English. As discussed before, we assume the domain data $D$ is given for each of the puzzles. A set of training puzzles, $\{P_1,...,P_n\}$ is used to train the natural language model which can be used translate natural language sentences into their ASP representations. This model is then used to translate clues for new puzzles. The initial dictionary contained nouns with most verbs. A set of testing puzzles, $\{P'_1,...,P'_m\}$, is validated by transforming the data into the proper format, adding generic modules and translating the clues of $P'_1,...,P'_m$ using the trained model. 

To evaluate our approach, we considered $50$ different logic puzzles from various magazines, such as \cite{puzzles,puzzles1,puzzles2}. We focused on evaluating the accuracy with which the actual puzzle clues were translated. In addition, we also verified the number of puzzles we solved. Note that in order to completely solve a puzzle, all the clues have to be translated accurately, as a missing clue means there will be several possible answer sets, which in turn will give an exact solution to the puzzle. Thus if a system would correctly translate $90\%$ of the puzzle clues, and assuming the puzzles have on an average 10 clues, then one would expect the overall accuracy of the system to be $0.9^{10} = 0.349$, or around $34.9\%$. 

To evaluate the clue translation, $800$ clues were selected. Standard 10 fold cross validation was used. $Precision$ measures the number of correctly translated clues, save for permutations in the body of the rules, or head of disjunctive rules. $Recall$ measures the number of correct exact translations. 

To evaluate the puzzles, we used the following approach. A number of puzzles were selected and all their clues formed the training data for the natural language module. The training data was used to learn the meaning of words and the associated parameters and these were then used to translate the English clues to ASP. These were then combined with the corresponding puzzle domain data, and the generic/background ASP module. The resulting program was solved using $clingo$, an extension of $clasp$ \cite{clingo}. $Accuracy$ measured the number of correctly solved puzzles. A puzzle was considered correctly solved if it provided a single correct solution. If a rule provided by the clue translation from English into ASP was not syntactically correct, it was discarded. We did several experiments. Using the $50$ puzzles, we did a 10-fold cross validation to measure the accuracy. In addition, we did additional experiments with 10, 15 and 20 puzzle {\em manually chosen} as training data.  The manual choice was done with the intention to pick the training set that will entail the best training. In all cases, the $C\&C$ parser \cite{CCG} was used to obtain the syntactic parse tree.

\subsection{Results and Analysis}

The results are given in tables \ref{tab:r1} and \ref{tab:r2}. The ``10-fold'' corresponds to experiments with 10-fold validation, ``10-s'', ``15-s'' and ``20-s'' to experiments where 10, 15 and 20 puzzles were manually chosen as training data respectively.

\begin{minipage}[b]{.40\textwidth}
\centering
\begin{small}
\begin{tabular}{| c| c| c|}
\hline
 Precision & Recall & F-measure \\
\hline
  87.64 & 86.12 & 86.87 \\
\hline
\end{tabular}
\end{small}
\captionof{table}{Clue translation performance.}
\label{tab:r2}
\end{minipage}\qquad

\begin{minipage}[b]{.40\textwidth}
\centering
\begin{small}
\begin{tabular}{|c| c|}
\hline
 & Accuracy \\
\hline
10-Fold &  28/50 $(56\%)$ \\
\hline
10-s &  22/40 $(55\%)$ \\
\hline
15-s &  24/35 $(68.57\%)$ \\
\hline
20-s &  25/30 $(83.33\%)$ \\
\hline
\end{tabular}
\end{small}
\captionof{table}{Performance on puzzle solving.}
\label{tab:r1}
\end{minipage}\qquad

The results for clue translation to ASP is comparable to translating natural language sentences to Geoquery and Robocup domains used by us in \cite{me:iwcs}, and used in similar works  such as \cite{Collins:2007} and \cite{Mooney:2009}. Our results are close to the values reported there, which range from $88$ to $92$ percent for the database domain and $75$ to $82$ percent for the Robocup domain.  

%The results are within the expected performance. 
As discussed before, a  $90\%$ accuracy is expected to lead to around $35\%$ rate for the actual puzzles. Our result of $56\%$ is significantly higher. It is interesting to note that as the number of puzzles used for training increases, so does the accuracy. However, there seems to be a ceiling of around $83.3\%$.  

In general, the reason for not being able to solve a puzzle lies in the inability to correctly translate the clue. Incorrectly translated clues which are not syntactically correct are discarded, while for some clues the system is not capable to produce any ASP representation at all. There are several major reasons why the system fails to translate a clue. First, even with large amount of training data, some puzzles simply have a relatively unique clue. For example, for the clue, ``The person with Lucky Element Earth had their fortune told exactly two days after Philip.'' the ``exactly two days after'' part is very rare and a similar clue, which discusses the distance of elements on a time line is only present in two different puzzles. There were only 2 clues that contain ``aired within n days of each other'', both in a single puzzle. If this puzzle is part of the training set, since we are not validating against it, it has no impact on the results. If it's one of the tested puzzles, this clue will essentially never be translated properly and as such the puzzle will never be correctly solved. In general, many of the clues required to solve the puzzles are very specific, and even with the addition of generic knowledge modules, the system is simply not capable to figure them out. A solution to this problem might be to use more background knowledge and a larger training sample, or a specific training sample which focuses on various different types of clues. In addition, when looking at tables \ref{tab:ex1} and \ref{tab:ASPparres}, many of the words are assigned very simple semantics that essentially do not contribute any meaning to the actual translation of the clue. Compared to database query language and robocup domains, there are several times as many simple representations. This leads to several problems. One of the problems is that the remaining semantics might be over fit to the particular training sentences. For example, for ``aired within n days of each other'' the only words with non trivial semantics might be ``within'' and some number ``n'', which in turn might not be generic for other sentences. The generalization approach adopted from \cite{me:iwcs} is unable to overcome this problem. The second problem is that a lot of words have these trivial semantics attached, even though they also have several other non trivial representations. This causes problem with learning, and the trivial semantics may be chosen over the non-trivial one. Finally, some of the $C\&C$ parses do not allow the proper use of inverse $\lambda$ operators, or their use leads to very complex expressions with several applications of $@$. In table \ref{tab:ex1}, this can be seen by looking the representation of the word ``immediately''. While this particular case does not cause serious issues, it illustrates that when present several times in a sentence, the resulting $\lambda$ expression can get very complex leading to third or fourth order $\lambda$-ASP-calculus formulas.

\section{Conclusion and Future work}

In this work we presented a learning approach to solve combinatorial logic puzzles in English. Our system uses an initial dictionary and general knowledge modules to obtain an ASP program whose unique answer set corresponded to the solution of the puzzle. Using a set of puzzles and their clues to train a model which can translate English sentences into logical form, we were able to solve many additional puzzles by automatically translating their clues, given in simplified English, into ASP.  Our system used results and components from various AI sub-disciplines including natural language processing, knowledge representation and reasoning, machine learning and ontologies as well as the functional programming concept of $\lambda$-calculus. There are many ways to extend our work. The simplified English limitation might be lifted by better natural language processing tools and additional sentence analysis. We could also apply our approach to different types of puzzles. A modified encodings might yield a smaller variance in the results.  Finally we would like to submit that solving puzzles given in a natural language could be considered as a challenge problem for human level intelligence as it encompasses various facets of intelligence that we listed earlier. In particular, one has to use a reasoning system and can not substitute it with surface level analysis often used in information retrieval based methods.

\bibliographystyle{aaai}
\bibliography{bib-all}

\end{document}